\documentclass[11pt,letterpaper]{llncs}
\usepackage{amsmath}
\usepackage{amssymb}
\usepackage{graphicx}
\usepackage{marvosym}
\usepackage{url}
\usepackage{fancyhdr}

\usepackage{etoolbox}
\usepackage{setspace}
\AtBeginEnvironment{quote}{\singlespacing} 
\AtEndEnvironment{quote}{\singlespacing}

\usepackage{geometry}
\newcommand{\keywords}[1]{\par\addvspace\baselineskip
\noindent\keywordname\enspace\ignorespaces#1}

\fancypagestyle{firstpage}{\fancyhf{}
\setcounter{page}{1}
\fancyhead[C]{\small{Accepted paper in 4th International Conference on Semantic \& Natural Language Processing (SNLP 2023)}}
\rfoot{\thepage}
}

\begin{document}


\title{\LARGE{Question-type Identification for Academic Questions in Online Learning Platform}}


%
%

\author{\large{Azam Rabiee, Alok Goel, Johnson D’Souza, Saurabh Khanwalkar}}
\institute{\large{Course Hero, Inc.}}

%


%
%


\maketitle

\thispagestyle{firstpage}

\begin{abstract}
\textit{Online learning platforms provide learning materials and answers to students' academic questions by experts, peers, or systems. This paper explores question-type identification as a step in content understanding for an online learning platform. The aim of the question-type identifier is to categorize question types based on their structure and complexity, using the question text, subject, and structural features. We have defined twelve question-type classes, including Multiple-Choice Question (MCQ), essay, and others. We have compiled an internal dataset of students' questions and used a combination of weak-supervision techniques and manual annotation. We then trained a BERT-based ensemble model on this dataset and evaluated this model on a separate human-labeled test set. Our experiments yielded an F1-score of 0.94 for MCQ binary classification and promising results for 12-class multilabel classification. We deployed the model in our online learning platform as a crucial enabler for content understanding to enhance the student learning experience.}

\end{abstract}


\section{Introduction}

The very idea of adaptive and personalized learning comes from the understanding that students have individual learning needs based on their curriculum and level of schooling in addition to their aptitudes, learning styles, and environments, among other significant factors. Machine learning practitioners can develop a deep understanding of students’ academic questions and match them with the optimal content that will allow users to make the best decisions about how to drive their unique learning journey forward. Over time, these personalized learning engines will learn how students learn best, what concepts to focus on, and the temporal relevance of content, eventually becoming a proactive and assistive educational tool for every learner, regardless of background, school, or personal learning style. 

\begin{figure}
\centering
\includegraphics[height=3in, width=3in]{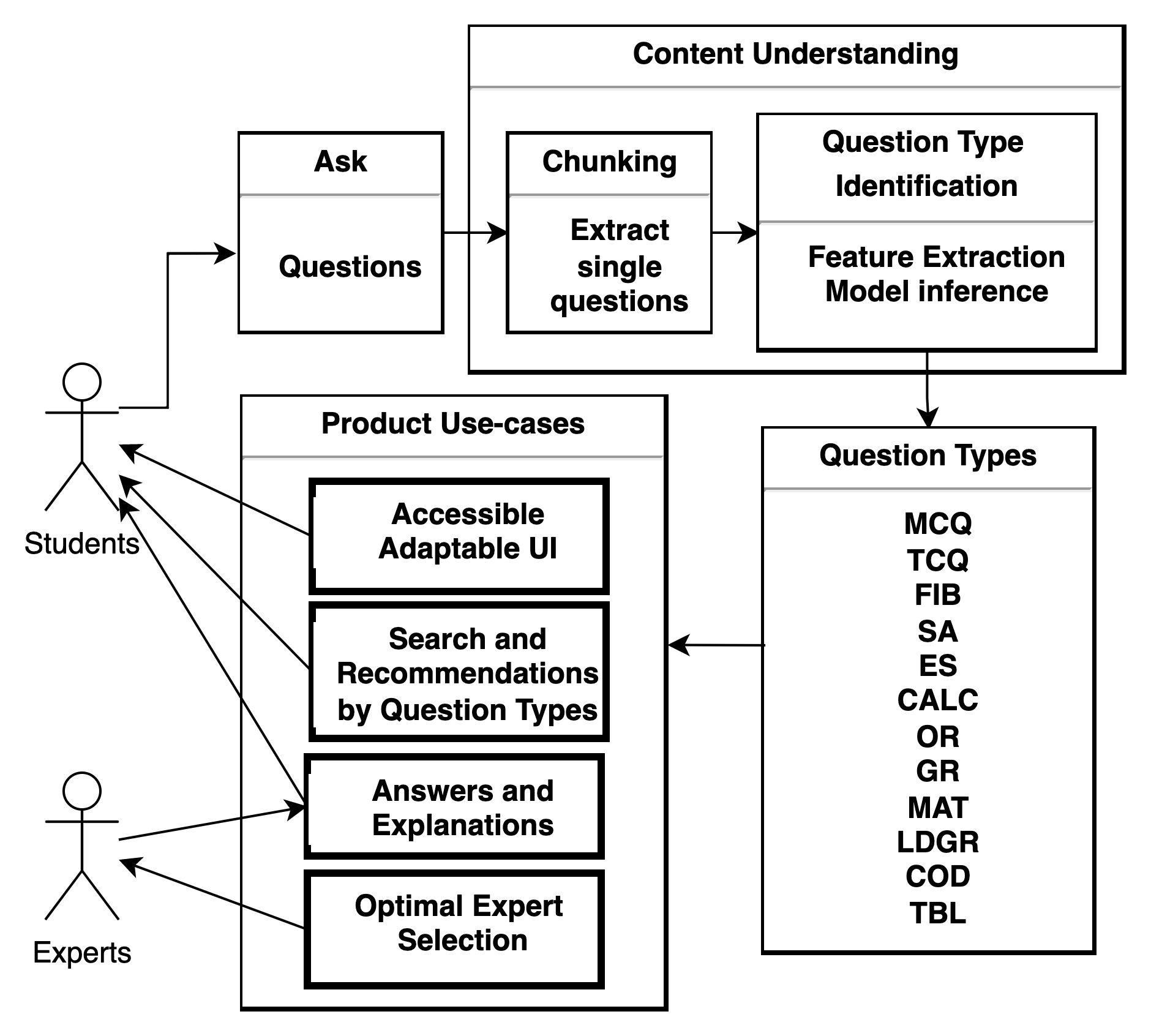}
\caption{Question-type identification for a personalized learning engine}
\label{fig:main-diag}
\end{figure}

A content understanding system is the cornerstone of this ubiquitous and universal learning experience. It enables personalized learning by attempting to serve every student's learning needs across subjects and knowledge levels. The content understanding system comprises an ensemble of machine learning models that segment student questions and study resources. One such method of segmentation is a question-type identifier. 
Figure~\ref{fig:main-diag} shows an example question flow overview in personalized learning engines. Students ask various types of questions. As illustrated in Figure~\ref{fig:main-diag}, the questions first go through a content understanding system, which includes chunking to separate individual questions. Then, each question is passed to the \emph{Question Type Identification} system, which includes feature extraction and model inference. The identifier provides one or more question-type classes. The identified question-types can be used in various product use-cases, including accessible and adaptable UI, search and recommendations, answers and explanations, and optimal expert selection. 

Identifying question types and how they serve different user intents can serve as a personalization lever for helping students understand the underlying concepts better. In addition, the personalized learning engine can utilize the question types for improving question discovery and recommendations. For example, multiple choice questions (MCQs) may seem more actionable for a student intending to prepare for an exam \cite{einig2013:supporting}, while short answer questions facilitate conceptual learning for completing homework assignments and essays.

The use cases of the identifier are not limited to the above-mentioned items. Having the feature in our personalized learning engine enables us to built an infrastructure for smart content understanding with a wide range of functionalities. Understanding the different types of questions allows us to help students in the following ways:

\begin{enumerate}
  \item Enables students to access content seamlessly across applications by enabling user interface (UI) changes based on the question format and length.
  \item Improves students' ability to practice by understanding their immediate learning objective and matching content recommendations accordingly.
  \item Improves concept explanations by linking students with content from domain-specific knowledge sources.
  \item Optimally routes students to expert help based on the complexity and concepts of their questions.
\end{enumerate}

Our ideas for the identifier have been implemented, evaluated, and deployed in our online learning platform, which contains course-specific study resources including user-submitted questions and study guides. The identifier was used to set different tutor pricing (the cost for tutors to answer a question) based on question type (MCQ and non-MCQ). A/B testing showed that the pricing change was successful without any statistically significant difference in answer rate, tutor response time, or students' helpfulness ratings. In an internal analysis of document usage by students on our platform, it was discovered that documents with answered MCQs are four times more valuable than documents with answered non-MCQs. This shows another emerging usage of the question identifier to prioritize answering some question-types over others. This would enable students to get more answers for the question-types that they care more about. The identifier can also be used to better understand students' and tutors' preferences in terms of question types, so that students get question recommendations based on question types, and tutors are assigned questions based on their past answers and ratings for the question types. Certain question-types are easier to answer than others; newer tutors can be assigned easier questions, and students can be recommended questions based on question type difficulty.
	
This paper will explore how to build such a question-type identifier with the following contributions: 
\begin{enumerate}
  \item There is no related work to our knowledge that uses machine learning to train a model for question identification that can be integrated into the real-world application of the online learning system.
  \item Lack of enough labeled data is one of the challenges of developing the identifier. Our experiments rely on data augmentation and a weak supervision technique (Snorkel \cite{ratner2020:snorkel}).  
  \item We introduced an ensemble and multi-modal architecture for this task, which, to our knowledge, is novel and gives demonstrably better results. For this purpose, we extracted handcrafted features alongside BERT model \cite{devlin2018:bert} outputs into another classifier (XGBoost \cite{chen2016:xgboost}). 
\end{enumerate}

The rest of the paper is organized in the following manner. Section~\ref{sec:related-work} reviews related work in order to clearly illustrate our contribution and development in existing studies. Section~\ref{sec:qti} demonstrates the question-type identifier definitions. We explain our dataset in Section~\ref{sec:data}. The experimental setup, model, and results of our identifier are discussed in Section~\ref{sec:exp}. Finally, Section~\ref{sec:con} offers some concluding remarks.

\section{Related Work}
\label{sec:related-work}
Adaptive and personalized learning has emerged as a fundamental learning paradigm in the educational technology research community \cite{shemshack2020:systematic}. Machine learning approaches also enable the personalization of content for a wide diversity of students \cite{StHilaire2022:ANE}. For example, Hypocampus \cite{lincke2021:performance} is a personalized learning engine for children and youth relying on the hypothesis that introducing time intervals between study sessions using a \emph{repetitive rule} can improve memory and boost recall. It introduced an adaptive repetitive rule for knowledge retention in the long-term memory by analyzing students’ learning style, study schedule, knowledge, and performance. Khosravi et al. in \cite{khosravi2017:riple} studied another aspect of a personalized learning engine for helping students to find resources based on their interests and knowledge gaps. These learning resource recommendations, however, are limited to a few specific subjects \cite{zhou2021:resource} and do not take into account question types.  

Question-type identification in contemporary research literature focuses on the following broad application categories:
\begin{enumerate}
  \item Improved Personalization
  \item Improved Content Understanding
  \item Optimal Expert Selection
\end{enumerate}
 Sections~\ref{sec:imp-per}, \ref{sec:content}, and \ref{sec:tutor} review studies aligned with the above three categories. Then, Section~\ref{sec:qti-rev} surveys technical approaches of question-type identification. 

\subsection{Improved Personalization}
\label{sec:imp-per}
Students' perception of question types has a significant impact on their approaches to learning. This hypothesis has been well studied in \cite{holzinger2020:assess}, where the authors concluded that students preferred short answer questions (SAQs) for achieving long-term knowledge but preferred the objective format of MCQs for assessing their knowledge. Similarly, measurement of higher cognitive skills in undergraduate students has shown that MCQs perform better in student problem-based learning setups compared to Modified Essay Questions (MEQs) \cite{moeen2011:meq}. Question-types have also been instrumental in measuring the objective performance of student cohorts. High-achieving and low-achieving students consistently perform well in SAQs, followed by MCQs and MEQs \cite{bodkha2012:effectiveness}. Additionally, students with specific learning difficulties find it challenging to put their thoughts succinctly into essays. Question-types, specifically MCQs, have been shown to play an impactful role in mitigating those challenges \cite{photopoulos2021:preference}.

\subsection{Improved Content Understanding}
\label{sec:content}
Question classification has been extensively studied in question answering systems (QASs) as there is a significant link between the said question type and the expected answer type \cite{bouziane2015:question}. These QA systems target question classification tasks in both the open domain \cite{xu2020:multi,talmor2019:commonsenseqa,mohasseb2018:classification,mohasseb2018:question} and closed domain question answering  \cite{pechsiri2016:developing,kearns2018:resource,cruchet2008:supervised}. A lot of the focus here is on finding answers to questions quickly and efficiently by using extractive and reasoning-based methods that use semantic markers of the question. Li and Roth \cite{li2002:trec} have built popular question classification taxonomies and datasets that relate to the "aboutness" of questions, where the emphasis is on uncovering the meaning behind a question. Our novel work presented in this paper emphasizes not only the semantic nature of the question text but, more importantly, the structure of the question. We hypothesize better learning outcomes for students by focusing on question structure-based applications, as discussed in Section~\ref{sec:imp-per}. Structure-based question classification has been partly studied in \cite{mohasseb2018:question} where in addition to grammar and syntactic features of the question, the authors have also used question and answer structure metadata like \emph{Confirmation} (answers are either Yes or No) and \emph{Choice} (Similar to MCQ) to inform their downstream User Intent classification tasks for a QA system.

\subsection{Optimal Expert Selection}
\label{sec:tutor}
Question classification is capable of helping users of QAS get answers to their questions promptly by routing user questions to systems with a high probability of providing the correct answer. These systems could be domain-specific search engines that use information retrieval (IR) approaches to find the correct answer  \cite{rajpurkar2018:know}, systems that use structured knowledge bases  \cite{perevalov2022:qald}, using crowdsourced \cite{zhou2012:classification} or expert-sourced answers. Historically, these approaches excluded question structure-related signals when routing questions to experts. However, by including the question type as a structural feature, it becomes increasingly possible to find the optimal expert with the right skill level to answer a given question.

\subsection{Technical Background}
\label{sec:qti-rev}
Question-type identification is a text classification task that was previously performed by applying typical classifiers such as maximum entropy, decision trees, or support vector machines to grammatical, domain-expert, and linguistic features \cite{mohasseb2018:question,huang2009:investigation,chernov2015:linguistically}. Recent studies utilize recurrent and convolutional neural networks (RNNs and CNNs) on the question text embeddings for the question classification \cite{kearns2018:resource,seidakhmetov2020:question}. While ELMo has been shown to outperform GloVe and Word2Vec for text embedding \cite{kearns2018:resource}, BERT has outperformed others in various NLP tasks \cite{devlin2018:bert}. In this paper, we investigate BERT-based architectures in an ensemble model that combins both domain expert and text embedding features for question-type identification.

\section{Question-type Identification}
\label{sec:qti}

Question-type identification aims to classify single- and multi-sentence students’ questions into predefined question-types, using the question text, subject, and structural features. We have defined twelve question-types, summarized in Table \ref{tab:qtypes}, appropriate for our content understanding use cases.

\begin{table}
\centering
\renewcommand{\arraystretch}{1}  
\caption{Summary of question-types}
\begin{tabular}{ll} \hline
\textbf{Q-Type} & \textbf{Description}\\ \hline
MCQ & Multiple Choice Question, Multi-select \\ 
TCQ & Two Choice Question, True/False \\ 
FIB & Fill In the Blank \\
SA & Short Answer, Factoid questions \\
ES & Essay, Subjective and open-ended questions \\
CALC & Calculation, Mathematical solution or proof \\
OR & Ordering \\
GR & Graphing, Drawing, Sketching \\
MAT & Matching, Classification \\
LDGR & Ledger \\
COD & Coding, Programming, Pseudo-code \\
TBL & Table completion, Form filling \\
\hline
 \end{tabular}
 \label{tab:qtypes}
\end{table}

Regardless of the answer, we have defined the question-types according to their syntactic patterns, structure, and complexity. For example, the following question is an essay question-type (ES) that is subjective and open-ended. 

\begin{quote}
\texttt{Discuss the benefits and constraints of different network types and standards.}
\end{quote}

Short Answer (SA) is the question-type that is akin to a factoid question and has a direct and widely accepted fact-based answer, independent of the length of the answer. In ordering questions (OR), arranging items in a desired order is asked, such as: 
\begin{quote}
\texttt{Arrange the following groups in order of increasing priority.} 
\end{quote}

Graphing questions (GR) need the sketching or drawing of a diagram, circuit, flowchart, or histogram. Matching (MAT) is another question type that asks for the classification of items or the matching of two lists. Accounting questions that require journalizing transactions and making balance sheets are defined as ledger question-type (LDGR). Code type (COD) covers all programming, coding, and debugging questions. Finally, table type (TBL) includes questions that require completing a table or filling out forms. 

\subsubsection{Multilabeling}

In our application, a single question can have multiple question-types associated with it. For example, if an MCQ includes calculation, it is labeled as both MCQ and CALC. It is because some question-types, such as MCQ, two choice question (TCQ) and fill in the blank (FIB), are purely structural and can be combined with each other and other types. Two examples are given in the following.

\textbf{Example 1:} MCQ and CALC
\begin{quote}

\texttt{We say that a three-digit number is balanced if the middle digit is the arithmetic mean of the other two digits. How many balanced numbers are divisible by 18? \\ (A) 2 \\(B) 3 \\(C) 6 \\(D) 9 \\(E) 18}
\end{quote}

\textbf{Example 2:} MCQ and FIB
\begin{quote}

\texttt{During the proliferative phase of wound healing, ... build new tissue by secreting ... to take the shape of the original tissue. \\
A. Fibroblasts, collagen \\
B. Platelets, collagen \\
C. Mast cells, histamine \\
D. Neutrophils, keratin}
\end{quote}

In addition, some questions are multi-part, with different question types in each part. For example, the question that asks for a calculation in part (a) and drawing a graph in part (b) is labeled as CALC and GR. With that in mind, in this paper, we have explored machine learning models for multilabel classification with one-versus-all classifiers.  

\subsubsection{Handcrafted Features}

We have found that some structural patterns and keywords play an important role in question-type identification. For example, the presence of an itemized list (like A., B., C., D.) or some MCQ keywords, such as "\texttt{Select all that apply}" and "\texttt{What is the correct answer,}" are reliable semantic features for the MCQ type. Similarly, "\texttt{Discuss strengths and weaknesses of...}" is one of the phrases related to the ES type. In this paper, we  refer to the features extracted according to the structural patterns and type-based keywords as \emph{handcrafted} features. For question-type identification, our classification approach takes into account handcrafted features as well as question text embedding. We explored the BERT \cite{devlin2018:bert} and DistilBERT \cite{sanh2019:distilbert} models to classify question text embeddings and the XGBoost algorithm \cite{chen2016:xgboost} for numeric modality. We discuss our experiments in Section~\ref{sec:exp}.

\section{Datasets}
\label{sec:data}

\begin{table}[b]
\centering
\renewcommand{\arraystretch}{1.3}  
\caption{Summary of Datasets}
\begin{tabular}{lccl} \hline
\textbf{Name} & \textbf{Size} & \textbf{Question Types} & \textbf{Source} \\ \hline
Test set & 1k & All & Internal, manually annotated \\
Gold set & $<$ 1k & All & Internal, manually annotated \\
Answer-type set & 500k & 10 labels & Internal, labeled according to answers' format \\
Silver set & 25k & All & Internal, weak-labeled by Snorkel \cite{ratner2020:snorkel}\\
Augmented set & 15k & All & Internal, weak-labeled by semantic similarity\\
TREC \cite{li2002:trec} & 5.5k & SA & public \\
\hline
 \end{tabular}
 \label{tab:datasets}
\end{table}

We utilized both proprietary data and a public dataset. Our internal datasets contain students' questions that were either manually annotated or supervised using a weak supervision technique. The summary of our data is reported in Table~\ref{tab:datasets}. 

Our \emph{test} and \emph{gold} sets contain student questions submitted through our online tutoring platform that are manually annotated. The students' questions span over a hundred academic subjects, from economics to literature, biology to history, accounting to psychology, and other college courses. We didn't use students' personal information in the development of the question-type identifier.

Our \emph{answer-type} set contains half a million questions for 14 subjects. This dataset used 10 prelabeled question types that differ from our type definitions in Table~\ref{tab:qtypes}. Instead, these questions were defined based on the answers’ format, which does not align with our use cases. The questions are single-labeled and imbalanced in terms of both types and subjects: 44\% were CALC, 41\% were SA, and the remaining 15\% were distributed across additional classes. Therefore, this dataset is inappropriate due to its limited and imbalanced subjects. As a result, we used the dataset for initial fine-tuning of large models before fine tuning with a superior dataset whose distributions matched the real-world use cases of content understanding.

We curated an internal dataset for questions over 100 academic subjects and labeled them using Snorkel \cite{ratner2020:snorkel}, as a weak supervision technique. We named the weakly supervised set of 25k students' questions as the \emph{silver} set, as listed in Table~\ref{tab:datasets}. Snorkel uses labeling functions that capture domain knowledge and resources, which can have unknown accuracy and correlations. The labeling functions' outputs are denoised using a machine learning approach to create much larger training sets much more quickly and at a lower cost than manual labeling. We compiled a multilabel Snorkel generator using \emph{LabelModel} in the Snorkel Python library on a development set of 500 manually annotated questions with 40 labeling functions found during the annotation. 

The \emph{augmented} set was created by weak-labeling questions that are similar to the gold set using the semantic similarity of the questions' text embeddings. We picked 15k unlabeled questions that had 80\% to 95\% similarity score with samples in the gold set and labeled them with the same question type as similar samples. We ensured that there is no overlap between the augmented set and the test set. 

We also included the TREC public dataset \cite{li2002:trec} in our experiments. TREC is a dataset for classifying types of questions with 6 semantical classes of Abbreviation, Entity, Description, Human, Location and Numeric. It provides questions that mostly consist of one sentence. We categorized the samples in the dataset as SA in our application since they are all factoid questions that imply short answers. We empirically found it useful because we did not have enough labeling function in the Snorkel weak-supervision for the SA class.

\section{Experiments}
\label{sec:exp}

We explain the experimental setup, scope, and network architecture in Section~\ref{sec:setup}. In Section~\ref{sec:mcq}, we present the first set of experiments for MCQ identification. Then, Section~\ref{sec:multiclass} discusses a multilabel classifier in a combined model for all twelve question types.

\subsection{Setup}
\label{sec:setup}

The scope of our experiments is limited to English language questions and assumes that the identifier acts on a single question. Thus, student requests containing multiple questions should be broken down into single questions. An example of the chunking approach is discussed in \cite{takechi2007:chunking}.

\subsubsection{BERT-based Ensemble Model}

\begin{figure}
\centering
\includegraphics [height=2in, width=2.2in] {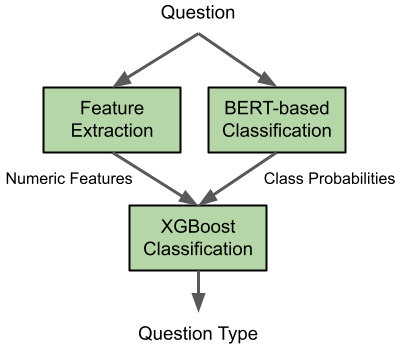}
\caption{A BERT-based ensemble model for question-type identification}
\label{fig:model}
\end{figure}

As we mentioned earlier, we explore BERT-based ensemble models in this study. Figure~\ref{fig:model} shows an example of a model that cascades a BERT-based architecture with another classifier. The former classifier provides the class probabilities according to the embedding of the question text. Later, the predicted class probabilities and a handcrafted feature vector are fed to the second classifier for the final decision. 

Manual annotation of our gold training set led to the design of the handcrafted features discussed earlier in Section~\ref{sec:qti}. Through the annotation process for making the ground-truth labeled dataset, we have created lists of numeric features related to each question-type. For example, a feature called \emph{count\_answer\_option} reflects the number of MCQ answer option keywords in the question text. The answer option keyword list contains 18 phrases such as "\texttt{None of the above}" or "\texttt{All of the above}." In total, we made a handcrafted 13-D feature vector for the MCQ classifier and a 175-D feature vector for the multilabel classification.

We have examined both BERT and DistilBERT in combination with the well-known XGBoost algorithm. We also compared the results with a Multimodal Transformer \cite{gu2021:package}, which similarly relies on a joint training procedure on the two modalities. The multimodal architecture utilizes BERT and an additional neural network (NN) architecture to support text as well as numerical and categorical features.

\subsection{MCQ Identifier}
\label{sec:mcq}
The first experiment focused only on identifying MCQs using BERT-based models. We consider MCQ and non-MCQ the highest order question-type categories that have an imbalanced distribution in our online tutoring platform. If a question has multiple types, and at least one of them is an MCQ, it will be considered an MCQ. For example, even if an MCQ requires calculations to find the correct answer option, it is still considered an MCQ despite CALC itself being a question-type. 

\subsubsection{Model Training}

We started both BERT and DistilBERT networks from HuggingFace BERT Sequence Classification pre-trained models \cite{wolf2020:transformers}. Then, we fine-tuned the models on a training set consisting of 400k samples of answer-type data and less than 1k manually annotated samples. Training the BERT model on an AWS SageMaker GPU (ml.p3.2xlarge) took around 14 hours with a batch size of 8 per device. Both models were trained on the first 512 word-piece tokens of the question text, converted to lowercase, and tokenized with the HuggingFace provided library.  

The MCQ probabilities predicted by the BERT and DistilBERT models were used as an input feature, along with the 13-D handcrafted feature vector, to train an XGBoost classifier model. The same gold and answer-type sets are used in training the XGBoost classifier. 
\emph{Hyperopt} library for Bayesian optimization \cite{bergstra2013:making} was utilized for hyperparameter tuning with Recall@Precision = 0.95 as the optimizer loss function. Finally, the classification probability predicted by the XGBoost model was compared against a threshold to provide MCQ or non-MCQ classification. 

\subsubsection{Results}

\begin{figure}
\centering
\includegraphics[height=3in, width=3.5in]{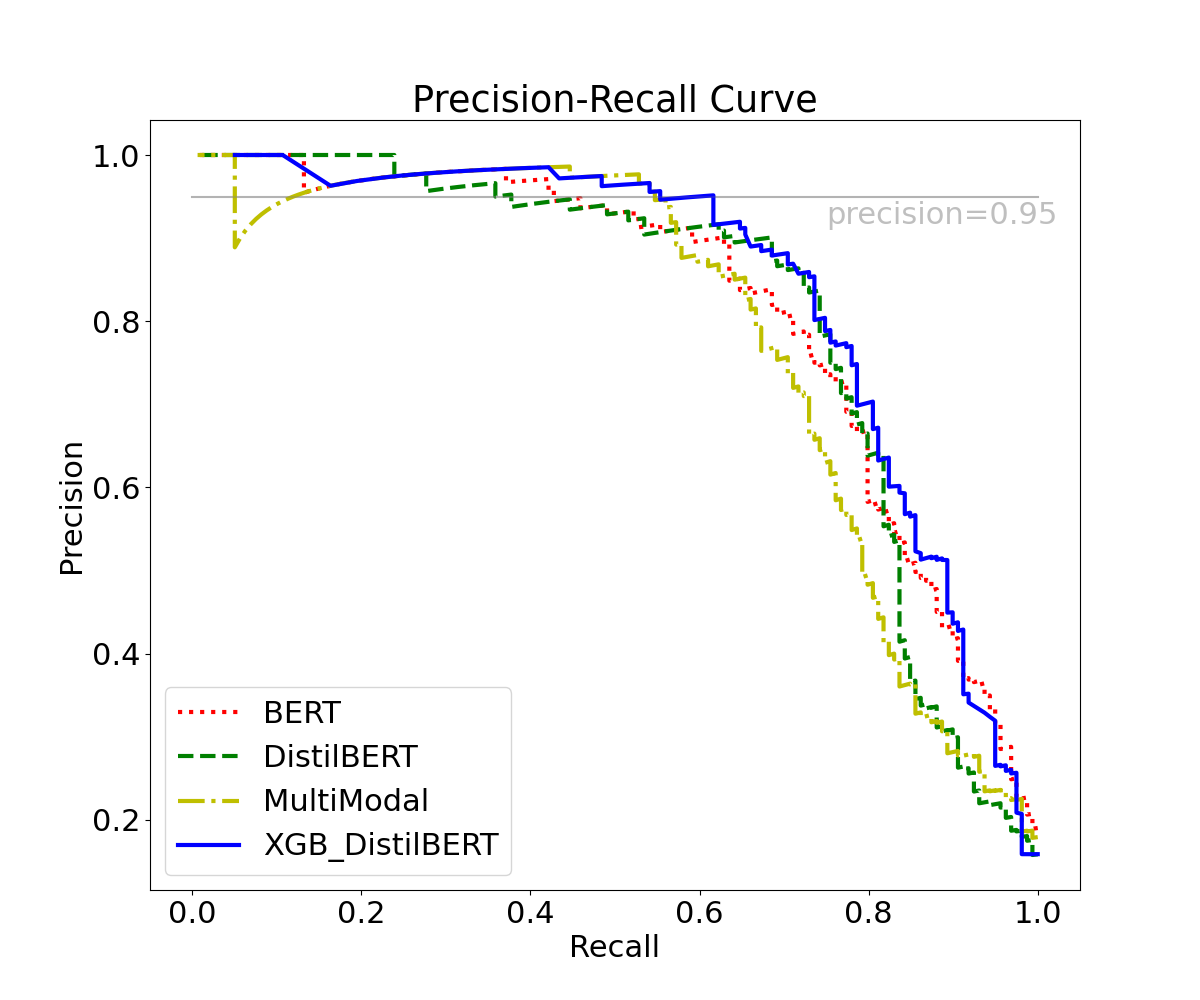}
\caption{Precision-Recall (PR) curve comparison}
\label{fig:pr}
\end{figure}

Table~\ref{tbl:mcq-res} reports the results in terms of accuracy (ACC), F0.5 (to weight precision higher than recall), weighted F1 (WF1, to consider each class’s support), and recalls at three different precisions (R@95, R@90, and R@85). F0.5 is calculated only for the MCQ class. For ACC, F0.5, and WF1, the classification threshold is 0.5; so if the prediction probability is higher than 0.5, the sample is considered MCQ. For R@N, we select the best threshold according to the PR curves. Figure~\ref{fig:pr} compares PR curves for BERT, DistilBERT, MultiModal, and XGB\_DistilBERT in detail.

\begin{table}
\centering
\renewcommand{\arraystretch}{1}  
\caption{Results of MCQ identifier on the test set}
\begin{tabular}{l|ccc|ccc} \hline
\textbf{Models} & \textbf{Acc} & \textbf{F0.5} & \textbf{WF1} & \textbf{R@95} & \textbf{R@90} & \textbf{R@85} \\ \hline
BERT & 93\% & 0.81 & 0.92 & 0.43 & 0.63 & 0.65 \\
DistilBERT & \textbf{94\%} & 0.84 & 0.93 & 0.38 & 0.69 & 0.72 \\ \hline
XGB with BERT & 93\% & 0.78 & 0.93 & 0.47 & 0.65 & \textbf{0.74} \\
XGB with DistilBERT & \textbf{94\%} & \textbf{0.85} & \textbf{0.94} & \textbf{0.62} & 0.65 & \textbf{0.74} \\ \hline
Multimodal Transformer & 93\% & 0.80 & 0.92 & 0.55 & 0.57 & 0.64 \\
\hline
\end{tabular}
\label{tbl:mcq-res}
\end{table}

As shown in Table~\ref{tbl:mcq-res} and Figure~\ref{fig:pr}, empirically, we found that DistilBert often performed better than BERT; this could be because the larger BERT models require more samples to train or because they would overfit the train set. The train set performance was significantly better than the validation and test set. To avoid overfitting, we leveraged different training parameters—specifically, changing the weight decay (L2 regularization) \cite{goodfellow2016:deep} found that some of the higher regularization values do improve test set scores. 

On an AWS SageMaker GPU (ml.p3.8xlarge), prediction latency for each student question is roughly 100 ms for the BERT model and even slower for the multimodal architecture. The prediction for the DistilBERT model takes half the time with the same settings. The final ensemble XGB\_DistilBERT model predicts the question-type almost as fast as the DistilBERT as the added XGBoost architecture takes only 0.01 ms to generate predictions.

In addition to the slow inference time, our experiments with the multimodal architecture were dissatisfying, and we ran into memory issues using the available library. However, the NN architecture for numerical and categorical features is not customizable to support different architectures. We only had control over how to combine scores from the BERT model and the added NN nodes.

\subsubsection{Error Analysis on MCQ Classifier}
To determine the reasons for low \emph{Recall} in our experiments, we manually analyzed the false-negative samples in the test set. We discovered that the question text is noisy for roughly 50\% of false-negative errors, meaning that either the question text is incomplete or the answer option symbol is missing. In addition, for around 70\% of false negatives, the question text does not contain any MCQ-related keyword. Here are two examples where our model could not predict as MCQ:

\textbf{Example 1:} non-MCQ predicted
\begin{quote}
\texttt{If a country has a birthrate of 13 and a death rate of 9, what is the \% increase? \\ Options: .2 .4 2 4}
\end{quote}

\textbf{Example 2:} non-MCQ predicted
\begin{quote}
\texttt{Systems of equations with different slopes and different y-intercepts have more than one solution. (5 points)   Always   Sometimes   Never}
\end{quote}

\subsection{Multilabel Classifier}
\label{sec:multiclass}

\begin{figure}
\centering
\includegraphics[height=2in, width=2.2in]{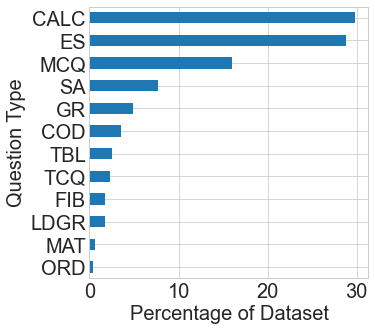}
\caption{Distribution of question-types asked by students in our online tutoring platform}
\label{fig:distribution}
\end{figure}

We trained a BERT-based ensemble model in a multilabel classification scheme for all twelve question-types. Figure~\ref{fig:distribution} depicts the distribution of question-types asked by students on our online tutoring platform, which is imbalanced and makes classification difficult. The figure shows that they ask more CALC, ES, and MCQ than ORD, MAT, and LDGR questions. 

\subsubsection{Model Training}

Similar to the previous experiments, we trained the BERT model from HuggingFace BERT Sequence Classification pre-trained models \cite{wolf2020:transformers} on a training set consisting of a golden set of 500 samples, answer-type, and TREC datasets \cite{li2002:trec}. Later, we fine-tuned the trained BERT model on the same gold set, the silver set, and the augmented gold set. 

To build the handcraft and numeric features for this experiment, we utilized the question subject and text because some question-types are more prevalent in specific educational domains. For example, most questions asked in the "Python programming" subject were identified as COD.

We trained the XGBoost model on the handcrafted features extracted from the same gold, answer-type, silver, and augmented sets. The TREC dataset is not used to train XGBoost since all samples are categorized as SA and this question-type is hardly identified with structural patterns or keywords for the handcrafted features. 

To get a multilabel prediction from the XGBoost model, we ran a one-vs-all experiment using the scikit-learn Python package \cite{ped2011:scikit-learn}. If no label was selected, we picked the label with the highest probability for the multilabel prediction. Similar to the previous experiments, we performed hyper-parameter tuning for the XGBoost model with F1-score as the optimizer loss function using \emph{Hyperopt} library \cite{bergstra2013:making}. The final inference output was obtained by averaging the probabilities of the BERT and XGBoost models. 

Training the BERT model on an AWS SageMaker GPU (ml.p3.2xlarge) took around 14 hours for the first fine-tuning and around 1.5 hours for the final retuning, with batch-size equals 8 per device. The prediction latency for the multilabel classifier with the ensembled XGBoost and retuned BERT model took roughly 100 ms on an AWS SageMaker GPU (ml.p3.8xlarge).

\subsubsection{Results}

\begin{table}[t]
\centering
\renewcommand{\arraystretch}{1}  
\caption{Test results of multilabel classification for twelve question types}
\begin{tabular}{lcc|ccc} \hline
\textbf{Models} & \textbf{WF0.5} & \textbf{WF1} & \textbf{F1\_CALC} & \textbf{F1\_ES} & \textbf{F1\_MCQ} \\ \hline
Snorkel Generator & 0.67 & 0.48  & 0.76 & 0.26 & 0.58 \\
BERT 1 (Gold, Answer-type, TREC) & 0.66 & 0.51  & 0.67 & 0.41 & \textbf{0.79} \\
BERT 2 (Gold, Silver, Augmented) & 0.69 & 0.63 & 0.80 & 0.75 & 0.54 \\
XGB & 0.66 & 0.61 & 0.67 & 0.71 & 0.57 \\
Ensembled BERT 2 and XGB  & \textbf{0.71} & \textbf{0.67} & \textbf{0.84} & \textbf{0.81} & 0.54 \\

\hline
 \end{tabular}
 \label{tbl:multi-res}
\end{table}

Table~\ref{tbl:multi-res} reports the experimental results in terms of weighted F0.5 and weighted F1 on the test set. We compared the final ensemble model against each individual model and the Snorkel Generator as a baseline model. We used the same Snorkel generator to create the silver set for our training data. As shown in the table, the final combined model improves the performance of each individual model. Furthermore, Table~\ref{tbl:multi-res} lists F1-scores for three top-frequent question-types. We have found that question types that are frequent in our dataset (CALC, ES, and MCQ) show high classification scores; but results are not yet satisfactory for other question-types. 

BERT 1 in Table~\ref{tbl:multi-res} refers to the first fine-tuning on gold, answer-type, and TREC sets. BERT 2 refers to the second fine-tuning on gold, silver, and augmented sets. Since our answer-type dataset contains only 10 classes (including MCQ), the overall performance of BERT 2 is higher than BERT 1. However, BERT 1 performed significantly better in MCQ.   

\subsubsection{Error Analysis on Multilabel Classifier}
By analyzing the results, we have found that our multilabel classifier has difficulties classifying ES and SA, which is a complex case even for humans. For example, the following question, which is defined as ES, is predicted as SA.
\begin{quote}
\texttt{How do we ensure we use the correct tRNA during translation?}  
\end{quote}

Another example of misclassification is the following question that is asked in the MySQL course and is labeled as COD but is predicted as SA.
\begin{quote}
\texttt{
Return the id, host name, and host id of the accommodation that has the highest "host\_total\_listings\_count". Return only the first result.}
\end{quote}
We believe that another reason for the lower performance in some classes is because of the limited manual annotations and the imbalance in question-types. Although the results are promising as a first step in building the multilabel classifier, we hope to continue improving the model for all 12 types. 

\section{Conclusions}
\label{sec:con}
We hypothesized that both the semantics and structure of a question are vital to gaining a deeper understanding of its type which makes it a multimodal classification problem. We investigated multiple ensemble models for MCQ identification and multilabel classification across 12 question types. We examined BERT and DistilBERT models for question-type classification on the question text embeddings and an XGBoost classifier to decide the numeric and handcrafted features extracted from the question text and subject. For a binary classification of MCQ versus non-MCQ, we also compared our results with multimodal transformers. To train the models, we used both internal and public datasets that were labeled manually or using weak supervision techniques. Our ensemble model improved the performance of each classifier independently. On a manually annotated test set, MCQ and 12-label classifiers showed an F1-score of 0.94 and 0.67, respectively.  

As an emerging feature, we deployed the question type identifier in our tutoring platform to assist students across numerous academic domains. The initial analysis and A/B testing convinced us of the usefulness of the identifier in improving content understanding and optimizing the routing of questions to the proper experts who can best support students when they are stuck. The smart content understanding enables a personalized learning engine to leverage MCQ, FIB, and TCQ questions to assist students with better practice. It can also utilize ES, SA, CALC, and other long-form question types to help students better understand critical concepts and topics.

These approaches for data generation and classification using both the semantics and structure of text can be used in various multimodal classification tasks, including query intent classification, document type classification, and other similar text classification problems.

\bibliography{QType_SNLP}

\begin{thebibliography}{10}
\providecommand{\url}[1]{#1}
\csname url@samestyle\endcsname
\providecommand{\newblock}{\relax}
\providecommand{\bibinfo}[2]{#2}
\providecommand{\BIBentrySTDinterwordspacing}{\spaceskip=0pt\relax}
\providecommand{\BIBentryALTinterwordstretchfactor}{4}
\providecommand{\BIBentryALTinterwordspacing}{\spaceskip=\fontdimen2\font plus
\BIBentryALTinterwordstretchfactor\fontdimen3\font minus
  \fontdimen4\font\relax}
\providecommand{\BIBforeignlanguage}[2]{{%
\expandafter\ifx\csname l@#1\endcsname\relax
\typeout{** WARNING: IEEEtran.bst: No hyphenation pattern has been}%
\typeout{** loaded for the language `#1'. Using the pattern for}%
\typeout{** the default language instead.}%
\else
\language=\csname l@#1\endcsname
\fi
#2}}
\providecommand{\BIBdecl}{\relax}
\BIBdecl

\bibitem{einig2013:supporting}
S.~Einig, ``Supporting students' learning: The use of formative online
  assessments,'' \emph{Accounting Education}, vol.~22, no.~5, pp. 425--444,
  2013.

\bibitem{ratner2020:snorkel}
A.~Ratner, S.~H. Bach, H.~Ehrenberg, J.~Fries, S.~Wu, and C.~R{\'e}, ``Snorkel:
  Rapid training data creation with weak supervision,'' \emph{The VLDB
  Journal}, vol.~29, no.~2, pp. 709--730, 2020.

\bibitem{devlin2018:bert}
J.~Devlin, M.-W. Chang, K.~Lee, and K.~Toutanova, ``Bert: Pre-training of deep
  bidirectional transformers for language understanding,'' \emph{arXiv preprint
  arXiv:1810.04805}, 2018.

\bibitem{chen2016:xgboost}
T.~Chen and C.~Guestrin, ``Xgboost: A scalable tree boosting system,'' in
  \emph{Proceedings of the 22nd acm sigkdd international conference on
  knowledge discovery and data mining}, 2016, pp. 785--794.

\bibitem{shemshack2020:systematic}
A.~Shemshack and J.~M. Spector, ``A systematic literature review of
  personalized learning terms,'' \emph{Smart Learning Environments}, vol.~7,
  no.~1, pp. 1--20, 2020.

\bibitem{StHilaire2022:ANE}
F.~St-Hilaire, D.~D. Vu, A.~Frau, N.~Burns, F.~Faraji, J.~Potochny, S.~Robert,
  A.~Roussel, S.~Zheng, T.~Glazier, J.~V. Romano, R.~Belfer, M.~Shayan,
  A.~Smofsky, T.~Delarosbil, S.~Ahn, S.~Eden-Walker, K.~Sony, A.~O. Ching,
  S.~Elkins, A.~Z. Stepanyan, A.~Matajov{\'a}, V.~Chen, H.~Sahraei, R.~Larson,
  N.~Markova, A.~Barkett, L.~Charlin, Y.~Bengio, I.~Serban, and E.~Kochmar, ``A
  new era: Intelligent tutoring systems will transform online learning for
  millions,'' \emph{arXiv preprint arXiv:2203.03724}, 2022.

\bibitem{lincke2021:performance}
A.~Lincke, M.~Jansen, M.~Milrad, and E.~Berge, ``The performance of some
  machine learning approaches and a rich context model in student answer
  prediction,'' \emph{Research and Practice in Technology Enhanced Learning},
  vol.~16, no.~1, pp. 1--16, 2021.

\bibitem{khosravi2017:riple}
H.~Khosravi, K.~Cooper, and K.~Kitto, ``Riple: Recommendation in peer-learning
  environments based on knowledge gaps and interests,'' \emph{arXiv preprint
  arXiv:1704.00556}, 2017.

\bibitem{zhou2021:resource}
Z.~Zhou, ``A resource recommendation algorithm for online english learning
  systems based on learning ability evaluation,'' \emph{International Journal
  of Emerging Technologies in Learning (iJET)}, vol.~16, no.~9, pp. 219--234,
  2021.

\bibitem{holzinger2020:assess}
A.~Holzinger, S.~Lettner, V.~Steiner-Hofbauer, and M.~Capan~Melser, ``How to
  assess? perceptions and preferences of undergraduate medical students
  concerning traditional assessment methods,'' \emph{BMC Medical Education},
  vol.~20, no.~1, pp. 1--7, 2020.

\bibitem{moeen2011:meq}
K.~Moeen-uz Zafar and M.~Badr-Aljarallah, ``Evaluation of modified essay
  questions (meq) and multiple choice questions (mcq) as a tool for assessing
  the cognitive skills of undergraduate medical students,'' \emph{International
  journal of health sciences}, vol.~5, no.~1, p.~39, 2011.

\bibitem{bodkha2012:effectiveness}
P.~Bodkha, ``Effectiveness of mcq, saq and meq in assessing cognitive domain
  among high and low achievers,'' \emph{IJRRMS}, vol.~2, no.~4, pp. 25--28,
  2012.

\bibitem{photopoulos2021:preference}
P.~Photopoulos, C.~Tsonos, I.~Stavrakas, and D.~Triantis, ``Preference for
  multiple choice and constructed response exams for engineering students with
  and without learning difficulties.'' in \emph{CSEDU (1)}, 2021, pp. 220--231.

\bibitem{bouziane2015:question}
A.~Bouziane, D.~Bouchiha, N.~Doumi, and M.~Malki, ``Question answering systems:
  survey and trends,'' \emph{Procedia Computer Science}, vol.~73, pp. 366--375,
  2015.

\bibitem{xu2020:multi}
D.~Xu, P.~Jansen, J.~Martin, Z.~Xie, V.~Yadav, H.~T. Madabushi, O.~Tafjord, and
  P.~Clark, ``Multi-class hierarchical question classification for multiple
  choice science exams,'' in \emph{Proceedings of the 12th Language Resources
  and Evaluation Conference}, 2020, pp. 5370--5382.

\bibitem{talmor2019:commonsenseqa}
A.~Talmor, J.~Herzig, N.~Lourie, and J.~Berant, ``Commonsenseqa: A question
  answering challenge targeting commonsense knowledge,'' in \emph{Proceedings
  of the 2019 Conference of the North American Chapter of the Association for
  Computational Linguistics: Human Language Technologies, Volume 1 (Long and
  Short Papers)}, 2019, pp. 4149--4158.

\bibitem{mohasseb2018:classification}
A.~Mohasseb, M.~Bader-El-Den, and M.~Cocea, ``Classification of factoid
  questions intent using grammatical features,'' \emph{ICT Express}, vol.~4,
  no.~4, pp. 239--242, 2018.

\bibitem{mohasseb2018:question}
------, ``Question categorization and classification using grammar based
  approach,'' \emph{Information Processing \& Management}, vol.~54, no.~6, pp.
  1228--1243, 2018.

\bibitem{pechsiri2016:developing}
C.~Pechsiri and R.~Piriyakul, ``Developing a why--how question answering system
  on community web boards with a causality graph including procedural
  knowledge,'' \emph{Information Processing in Agriculture}, vol.~3, no.~1, pp.
  36--53, 2016.

\bibitem{kearns2018:resource}
W.~R. Kearns and J.~A. Thomas, ``Resource and response type classification for
  consumer health question answering,'' in \emph{AMIA Annual Symposium
  Proceedings}, vol. 2018.\hskip 1em plus 0.5em minus 0.4em\relax American
  Medical Informatics Association, 2018, p. 634.

\bibitem{cruchet2008:supervised}
S.~Cruchet, A.~Gaudinat, and C.~Boyer, ``Supervised approach to recognize
  question type in a qa system for health,'' \emph{Studies in Health Technology
  and Informatics}, vol. 136, p. 407, 2008.

\bibitem{li2002:trec}
X.~Li and D.~Roth, ``Learning question classifiers,'' in \emph{COLING 2002: The
  19th International Conference on Computational Linguistics}, 2002.

\bibitem{rajpurkar2018:know}
P.~Rajpurkar, R.~Jia, and P.~Liang, ``Know what you don’t know: Unanswerable
  questions for squad,'' in \emph{Proceedings of the 56th Annual Meeting of the
  Association for Computational Linguistics (Volume 2: Short Papers)}, 2018,
  pp. 784--789.

\bibitem{perevalov2022:qald}
A.~Perevalov, D.~Diefenbach, R.~Usbeck, and A.~Both, ``Qald-9-plus: A
  multilingual dataset for question answering over dbpedia and wikidata
  translated by native speakers,'' \emph{arXiv preprint arXiv:2202.00120},
  2022.

\bibitem{zhou2012:classification}
T.~C. Zhou, M.~R. Lyu, and I.~King, ``A classification-based approach to
  question routing in community question answering,'' in \emph{Proceedings of
  the 21st international conference on world wide web}, 2012, pp. 783--790.

\bibitem{huang2009:investigation}
Z.~Huang, M.~Thint, and A.~Celikyilmaz, ``Investigation of question classifier
  in question answering,'' in \emph{Proceedings of the 2009 Conference on
  Empirical Methods in Natural Language Processing: Volume 2-Volume 2}, 2009,
  pp. 543--550.

\bibitem{chernov2015:linguistically}
A.~Chernov, V.~Petukhova, and D.~Klakow, ``Linguistically motivated question
  classification,'' in \emph{Proceedings of the 20th Nordic Conference of
  Computational Linguistics (NODALIDA 2015)}, 2015, pp. 51--59.

\bibitem{seidakhmetov2020:question}
T.~Seidakhmetov, ``Question type classification methods comparison,''
  \emph{arXiv preprint arXiv:2001.00571}, 2020.

\bibitem{sanh2019:distilbert}
V.~Sanh, L.~Debut, J.~Chaumond, and T.~Wolf, ``Distilbert, a distilled version
  of bert: smaller, faster, cheaper and lighter,'' \emph{arXiv preprint
  arXiv:1910.01108}, 2019.

\bibitem{takechi2007:chunking}
M.~Takechi, T.~Tokunaga, and Y.~Matsumoto, ``Chunking-based question type
  identification for multi-sentence queries,'' in \emph{Proceedings of the
  SIGIR 2007 Workshop on Focused Retrieval}, 2007, pp. 41--48.

\bibitem{gu2021:package}
K.~Gu and A.~Budhkar, ``A package for learning on tabular and text data with
  transformers,'' in \emph{Proceedings of the Third Workshop on Multimodal
  Artificial Intelligence}, 2021, pp. 69--73.

\bibitem{wolf2020:transformers}
T.~Wolf, L.~Debut, V.~Sanh, J.~Chaumond, C.~Delangue, A.~Moi, P.~Cistac,
  T.~Rault, R.~Louf, M.~Funtowicz \emph{et~al.}, ``Transformers:
  State-of-the-art natural language processing,'' in \emph{Proceedings of the
  2020 conference on empirical methods in natural language processing: system
  demonstrations}, 2020, pp. 38--45.

\bibitem{bergstra2013:making}
J.~Bergstra, D.~Yamins, and D.~Cox, ``Making a science of model search:
  Hyperparameter optimization in hundreds of dimensions for vision
  architectures,'' in \emph{International conference on machine
  learning}.\hskip 1em plus 0.5em minus 0.4em\relax PMLR, 2013, pp. 115--123.

\bibitem{goodfellow2016:deep}
I.~Goodfellow, Y.~Bengio, and A.~Courville, \emph{Deep learning}.\hskip 1em
  plus 0.5em minus 0.4em\relax MIT press, 2016.

\bibitem{ped2011:scikit-learn}
F.~Pedregosa, G.~Varoquaux, A.~Gramfort, V.~Michel, B.~Thirion, O.~Grisel,
  M.~Blondel, P.~Prettenhofer, R.~Weiss, V.~Dubourg, J.~Vanderplas, A.~Passos,
  D.~Cournapeau, M.~Brucher, M.~Perrot, and E.~Duchesnay, ``Scikit-learn:
  Machine learning in {P}ython,'' \emph{Journal of Machine Learning Research},
  vol.~12, pp. 2825--2830, 2011.

\end{thebibliography}
\bibliographystyle{IEEEtran}

\vspace{2cm}

\section*{Authors}
\noindent {\bf Azam Rabiee} received her PhD in Computer Engineering from Islamic Azad University, and she worked on Emotional Chatbots at KAIST, South Korea as a postdoc researcher. Currently, she is a Senior Machine Learning Engineer in Course Hero Inc. and her area of interests include Natural Language Processing, Speech Processing, and Time-series Predictions.\\

\noindent {\bf Alok Goel} received his Masters in Computer Science from North Carolina State University. He worked at IBM Watson as an Advisory Software Engineer on Knowledge Graphs and Discovery systems. Currently, he is a Senior Machine Learning Engineer at Course Hero. His interests include Natural Language Processing, Content Understanding, and Question Answering.\\

\noindent {\bf Johnson D’Souza} has over 16 years of industry experience and specializes in Natural Language Processing and Machine Learning for EdTech, Spoken Language Processing Systems and Consumer Electronics domains. He currently leads NLP initiatives at Course Hero with the goal of building delightful student experiences. Prior to Course Hero, Johnson worked at Verizon, DirecTV and TiVo.\\

\noindent {\bf Saurabh Khanwalkar} leads Course Hero's Machine Learning, Natural Language Processing, Search, and Recommendations teams and is responsible for shipping intelligent and personalized learning experiences to millions of students and educators. He has over 18 years of R\&D and leadership experience in machine learning products in diverse industries such as DARPA research, social media analytics, consumer electronics, healthtech, and edtech. Saurabh has technical publications and patents in Speech Processing, Natural Language Processing, and Information Retrieval and is on the review committee for NAACL, COLING, and other academic conferences offering industry track Machine Learning papers.

\end{document}